\documentclass[lettersize,journal]{IEEEtran}
\usepackage{amsmath,amsfonts}
\usepackage{algorithmic}
\usepackage{array}
\usepackage[caption=false,font=normalsize,labelfont=sf,textfont=sf]{subfig}
\usepackage{textcomp}
\usepackage{stfloats}
\usepackage{url}
\usepackage{verbatim}
\usepackage{graphicx}
\def\BibTeX{{\rm B\kern-.05em{\sc i\kern-.025em b}\kern-.08em
    T\kern-.1667em\lower.7ex\hbox{E}\kern-.125emX}}
\usepackage{balance}

\usepackage{times}
\usepackage{soul}
\usepackage{url}
\usepackage[hidelinks]{hyperref}
\usepackage[utf8]{inputenc}
\usepackage[small]{caption}
\usepackage{graphicx}
\usepackage{amsmath}
\usepackage{amsthm}
\usepackage{booktabs}
\usepackage{algorithm}
\usepackage{algorithmic}
\usepackage[switch]{lineno}
\usepackage{amsfonts,amssymb}
\usepackage{stfloats}
\usepackage{multirow}
\usepackage{graphicx}
\usepackage{cite}
\usepackage{cleveref}
\usepackage[numbers,sort&compress]{natbib}

\usepackage{colortbl}
\definecolor{light-gray}{gray}{0.82}
\newcolumntype{g}{>{\columncolor{light-gray}}c}

\usepackage{pifont}       
\usepackage{bbding}       
\usepackage{fontawesome}  

\usepackage{makecell}
 
\newcommand{\cmark}{\ding{51}}
\newcommand{\xmark}{\ding{55}}

\begin{document}
\title{NS-Net: Decoupling CLIP Semantic Information through NULL-Space for Generalizable AI-Generated Image Detection}
\author{Jiazhen Yan, Fan Wang, Ziwen He, Ziqiang Li, Zhangjie Fu,~\IEEEmembership{Member,~IEEE,}
\thanks{This work was supported in part by the National Natural Science Foundation of China under grant U22B2062, 62172232, T2600091, and Jiangsu Provincial Science and Technology Major Project (No. BG2024042). (Jiazhen Yan and Fan Wang contributed equally to this work.) (Corresponding author: Ziqiang Li, Zhangjie Fu).}
\thanks{Jiazhen Yan, Fan Wang, Ziqiang Li, Ziwen He and Zhangjie Fu are with the Engineering Research Center of Digital Forensics, Ministry of Education, Nanjing University of Information Science and Technology, Nanjing, 210044, China. (e-mail: 247918horizon@gmail.com, wf71103@126.com, iceli@mail.ustc.edu.cn, \{ziwen.he, fzj\}@nuist.edu.cn).}
}

\markboth{Journal of \LaTeX\ Class Files,~Vol.~18, No.~9, September~2020}%
{How to Use the IEEEtran \LaTeX \ Templates}

\maketitle

\begin{abstract}

The rapid progress of generative models, such as GANs and diffusion models, has facilitated the creation of highly realistic images, raising growing concerns over their misuse in security-sensitive domains. While existing detectors perform well under known generative settings, they often fail to generalize to unknown generative models, especially when semantic content between real and fake images is closely aligned. In this paper, we revisit the use of CLIP features for AI-generated image detection and uncover a critical limitation: the high-level semantic information embedded in CLIP's visual features hinders effective discrimination. To address this, we propose the NS-Net, a novel detection framework that leverages NULL-Space projection to decouple semantic information from CLIP's visual features, followed by contrastive learning to capture intrinsic distributional differences between real and generated images. Furthermore, we design a Patch Selection strategy to preserve fine-grained artifacts by selecting patches with the highest and lowest spectral entropy. Extensive experiments on an open-world benchmark comprising images generated by 47 diverse generative models show that NS-Net outperforms existing state-of-the-art methods, achieving a 5.0\% improvement in detection accuracy, thereby demonstrating strong generalization across both GAN- and diffusion-based image generation techniques. Our code is available at https://github.com/HorizonTEL/NS-Net.

\end{abstract}


\section{Introduction}
\label{sec:Introduction}
\IEEEPARstart{T}{he} rapid advancement of artificial intelligence has significantly accelerated the development of generative models, such as GANs \cite{karras2018progressive,huang2024gan}, diffusion models \cite{ho2020denoising,wu2024infinite}, and autoregressive models \cite{tian2024visual,tang2410hart}. While these models are widely employed to generate highly realistic images and have enabled numerous beneficial applications, their potential misuse raises serious concerns related to economic stability, political manipulation, and social security \cite{li2025prevent,yan2025detection,wang2025pair,wang2026map,hu2026high}.
Consequently, the reliable detection of AI-generated images has emerged as a critical research focus. To date, several effective detection methods \cite{wu2024local,tan2024frequency,tao2025sagnet,li2025pay} have been proposed. However, developing generalizable detection techniques that can effectively identify images from unknown generative models remains a fundamental challenge.

\begin{figure*}
    \centering
    \includegraphics[width=1\linewidth]{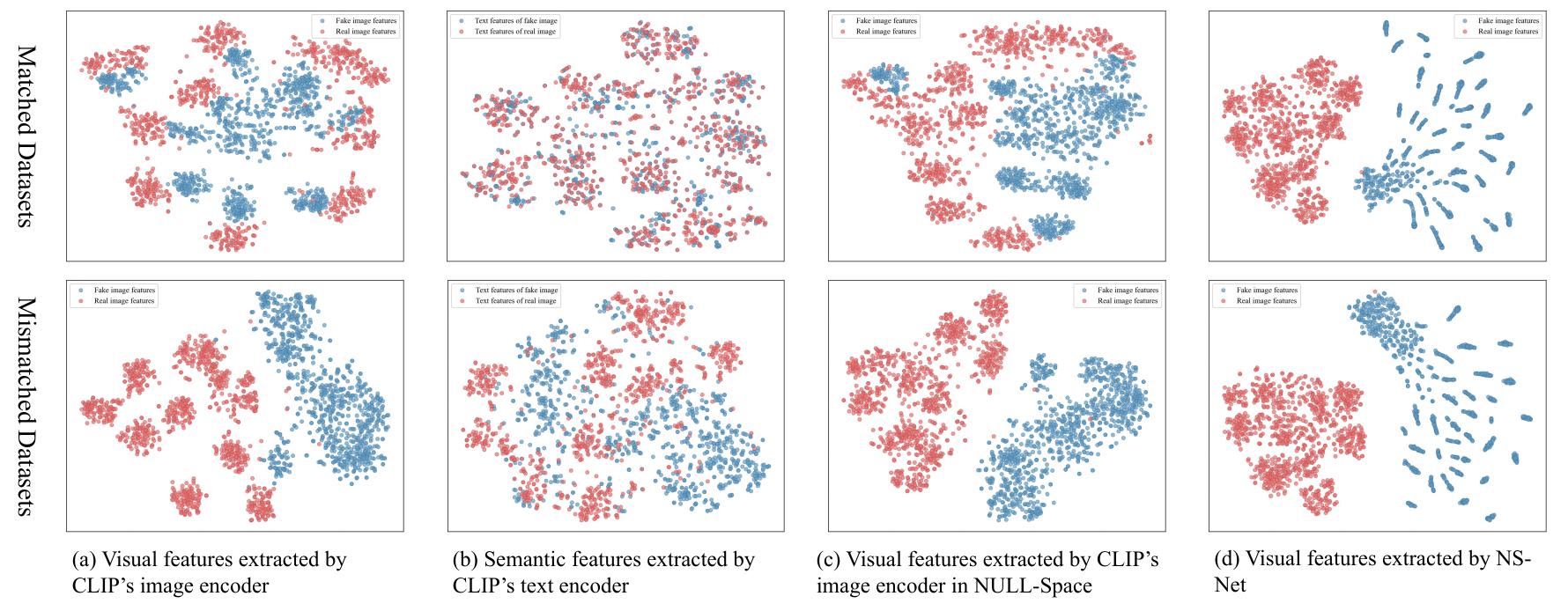}
    \caption{\textbf{T-SNE Visualization of Features Extracted from the Matched Dataset and the Mismatched Dataset}. The CLIP effectively separates real and generated images when semantic alignment is weak but struggles when the semantic content of both is closely aligned. Only applying the NULL-Space decoupling to features from the frozen CLIP’s image encoder results in a clear separation between real and generated images while reducing the influence of semantic content on forgery detection. }
    \label{fig:introduction}
\end{figure*}

Recently, a number of studies on AI-generated image detection \cite{ojha2023towards,liu2024forgery,fu2025exploring} have sought to enhance generalization by directly leveraging or fine-tuning CLIP \cite{radford2021learning}—a multimodal model pre-trained on large-scale image-text datasets, known for its strong generalization capabilities across unseen visual and textual data. Although the effectiveness of CLIP in this context has been demonstrated through quantitative experiments, there remains a lack of in-depth investigation into the intrinsic properties of its extracted features and how these characteristics influence detection performance.

Inspired by VIB-Net \cite{zhang2025towards}, which aims to decouple category-related features, we revisit the distinctions between features with different intrinsic properties extracted by CLIP. Specifically, we construct two datasets, each containing 1,000 real images from LSUN and 1,000 images generated by SDXL. In the \textit{matched dataset}, the generated images are conditioned on BLIP-derived textual descriptions of the corresponding real images. In contrast, the \textit{mismatched dataset} consists of randomly selected real images and generated images conditioned on unrelated text prompts. We extract visual features using CLIP's image encoder and visualize them using t-SNE, as shown in Figure~\ref{fig:introduction}(a). The features from the mismatched dataset exhibit clear clustering, suggestive of a linearly separable binary classification problem. In contrast, features from the matched dataset are more entangled and less distinguishable. This suggests that CLIP effectively separates real and generated images when semantic alignment is weak but struggles when the semantic content of both is closely aligned. Figure~\ref{fig:introduction}(b) presents the corresponding visualization of semantic features extracted by CLIP's text encoder. Based on these observations, we propose the following hypothesis: \textbf{The semantic information embedded in the visual features extracted by CLIP negatively impacts the performance of AI-generated image detection.}

To decouple the semantic information embedded in visual features, we construct the NULL-Space \cite{fang2024alphaedit} corresponding to the semantic components in CLIP’s image features. The NULL-Space, formally defined as the set of vectors mapped to the zero vector under a given linear transformation, enables the removal of specific feature components. Given that CLIP is trained for text-image alignment, we leverage text features as a more tractable and semantically explicit representation, rather than attempting to disentangle the complex semantics directly from visual features. Specifically, we extract features from the associated text prompts and compute their NULL-Space. Visual features extracted by CLIP’s image encoder are then projected onto this NULL-Space, thereby eliminating semantically aligned information and retaining components more relevant to forgery detection. As illustrated in Figure~\ref{fig:introduction}(c), only applying the NULL-Space decoupling to features from the frozen CLIP’s image encoder results in a clear separation between real and generated images while reducing the influence of semantic content on forgery detection. In addition, we also conduct an ImageNet-1K zero-shot classification. The original frozen CLIP image encoder achieves a Top-1 accuracy of 72.04\%. After applying NULL-Space decoupling to the visual features, the accuracy decreases to 15.36\%, further verifying that we have effectively suppressed a substantial amount of irrelevant semantic information in CLIP visual features.

To this end, we propose an innovative method, termed \textbf{NS-Net}, designed to extract more generalizable artifact information. Specifically, we leverage NULL-Space projection to decouple high-level semantic information embedded in CLIP's visual features, followed by contrastive learning to model the underlying distributional differences between real and generated images. This strategy improves generalization ability beyond simple classification accuracy. Moreover, the traditional CenterCrop \cite{li2025improving}, including PatchShuffle \cite{zhong2023patchcraft,zheng2024breaking}, only retains the artifacts at the center, overlooking critical forgery traces that may occur away from the center of the images. To mitigate this, we introduce a Patch Selection strategy. Each image is first divided into uniform patches, from which those with the highest and lowest entropy (as described in \cref{sec:patchselection}) are selected. These selected patches are then rearranged to form a new input image of the desired size, which maximizes the preservation of artifact information. As shown in Figure~\ref{fig:introduction}(d), our method effectively removes task-irrelevant semantic features while retaining the generator-specific artifact patterns crucial for reliable detection.

The main contributions are summarized as follows:

\begin{itemize}
\item We are the first to demonstrate that the semantic information embedded in the visual features extracted by CLIP adversely affects the performance of AI-generated image detection. To this end, we innovatively construct the NULL-Space of semantic features, which simply and efficiently eliminates task-irrelevant semantic information embedded in the image encoder’s visual representations, thus improving the generalization detection ability.

\item We design a data preprocessing strategy, Patch Selection, which maximizes the preservation of artifact information by selecting patches with the highest and lowest entropy.

\item Comprehensive experiments demonstrate that our method achieves superior generalization across 47 generative models, outperforming state-of-the-art method by 5.0\%.

\end{itemize}

\section{Related Works}
\label{sec:Related Works}

\subsection{AI-Generated Image Detection}
To mitigate the societal risks associated with AI-generated images, researchers have increasingly focused on developing effective detection methods. Recent research has shifted toward improving the generalization capabilities of detection models. Many of them focus on capturing \textbf{generalizable low-level forgery clues}, including local pixel relationships \cite{tan2024rethinking,cavia2024real}, frequency-based features \cite{tan2024frequency,yan2026dual}, gradients \cite{tan2023learning}, and reconstruction errors \cite{wang2023dire,chen2024drct}. To illustrate, F3-Net \cite{qian2020thinking} introduces frequency component division and the frequency statistical difference between real and forgery images into face forgery detection; NPR \cite{tan2024rethinking} reinterprets upsampling operations to construct an efficient artifact representation; DIRE \cite{wang2023dire} extracts the reconstruction error of the image over a pre-trained diffusion model as the artifact of diffusion-based images. In addition, some studies adopt \textbf{data-driven approaches} to enhance model generalization, focusing on data preprocessing \cite{wang2020cnn,chen2024drct,li2025improving} and the construction of augmented datasets \cite{guillaro2025bias,chen2025dual}, which improve the detection performance against sophisticated generation methods. For example, CNN-Spot \cite{wang2020cnn} employs diverse data augmentation techniques to enhance generalization to unseen testing data; SAFE \cite{li2025improving} integrates cropping and augmentations to improve generalization. What's more, B-Free \cite{guillaro2025bias} constructs debiased training samples via reconstruction, encouraging detectors to focus on generation-related artifacts rather than content, while DDA \cite{chen2025dual} proposes a dataset construction strategy that aligns pairs in both pixel and frequency domains to mitigate frequency-related bias.

In recent years, many researchers have explored leveraging CLIP \cite{radford2021learning} for detecting AI-generated images \cite{zhou2026breaking,zhou2026simplicity,yan2025dgs} and videos \cite{zheng2025d3}, owing to their powerful feature extraction and semantic understanding capabilities. UnivFD \cite{ojha2023towards} utilizes CLIP’s image features for binary classification, demonstrating promising generalization capabilities. However, subsequent studies \cite{liu2024forgery,fu2025exploring} reveal that CLIP’s native features are not fully optimized for forgery detection, prompting efforts to fine-tune the model to capture subtle artifacts. To this end, Fatformer \cite{liu2024forgery} introduces a forgery-aware adapter that integrates both spatial and frequency-domain extractors. C2P-CLIP \cite{tan2025c2p} further enhances CLIP’s detection capability by injecting category-consistent cues into the text encoder. CLIP-ADA \cite{deng2026clip} reveals that the mid-level stages primarily encode generator artifact features, and proposes the CLIP-guided Dual-level Augmentation and Forensic Distribution Adaptation framework to perform artifact-invariant generalizable detection. Additionally, VIB-Net \cite{zhang2025towards} addresses the issue of category interference in CLIP’s features by employing a Variational Information Bottleneck framework; Effort \cite{yan2024orthogonal} employs Singular Value Decomposition to construct two orthogonal subspaces to preserve the pre-trained knowledge while learning forgery-related patterns, achieving strong generalization performance.

Despite continuous advances, the intrinsic properties of CLIP's visual features remain insufficiently understood. We hypothesize that the entanglement of semantic information within these features impairs the model’s ability to capture task-specific artifact cues that are essential for forgery detection, motivating our goal to decouple the embedded semantic information in visual features.

\subsection{NULL-Space}
The NULL-Space \cite{coleman1986null} refers to the set of vector subspaces that are mapped to the zero vector under a linear transformation, and has been widely applied in natural language processing (NLP) \cite{ravfogel2020null,ravfogel2022linear}. In recent years, with the proliferation of large-scale pre-trained models, the concept of NULL-Space has also been increasingly adopted in continual learning \cite{zeng2019continual,guo2022adaptive,fang2024alphaedit} to mitigate catastrophic forgetting. Specifically, OWM \cite{zeng2019continual} projects new task updates onto the orthogonal complement of the past input subspace to prevent forgetting without replaying old samples, while AlphaEdit \cite{fang2024alphaedit} projects parameter perturbations onto the NULL-Space of knowledge to be retained, substantially reducing interference and forgetting during sequential editing.

As discussed above, projecting new knowledge into the NULL-Space of old knowledge makes it orthogonal to the old knowledge, thus mitigating catastrophic forgetting. By analogy, we treat semantic features as a subspace whose influence on visual representations we aim to eliminate. Accordingly, we project visual features onto the NULL-Space of this semantic subspace, ensuring that visual and semantic components are orthogonal. This orthogonality implies that variations in semantic features do not affect the projected visual representations, evidencing effective decoupling. Different from prior NULL-Space methods that preserve previously learned knowledge, we leverage NULL-Space to explicitly eliminate task-irrelevant semantics, yielding more generalized and artifact-relevant representations. Although both our method and Effort \cite{yan2024orthogonal} utilize SVD, they are fundamentally different in both core objectives and implementation details: Effort leverages SVD to derive an orthogonal subspace of pre-trained parameters to preserve semantic knowledge, while our method constructs NULL-Space of text features and explicitly removes task-irrelevant semantic information from visual features.

\begin{figure*}[!t]
    \centering
    \includegraphics[width=1\linewidth]{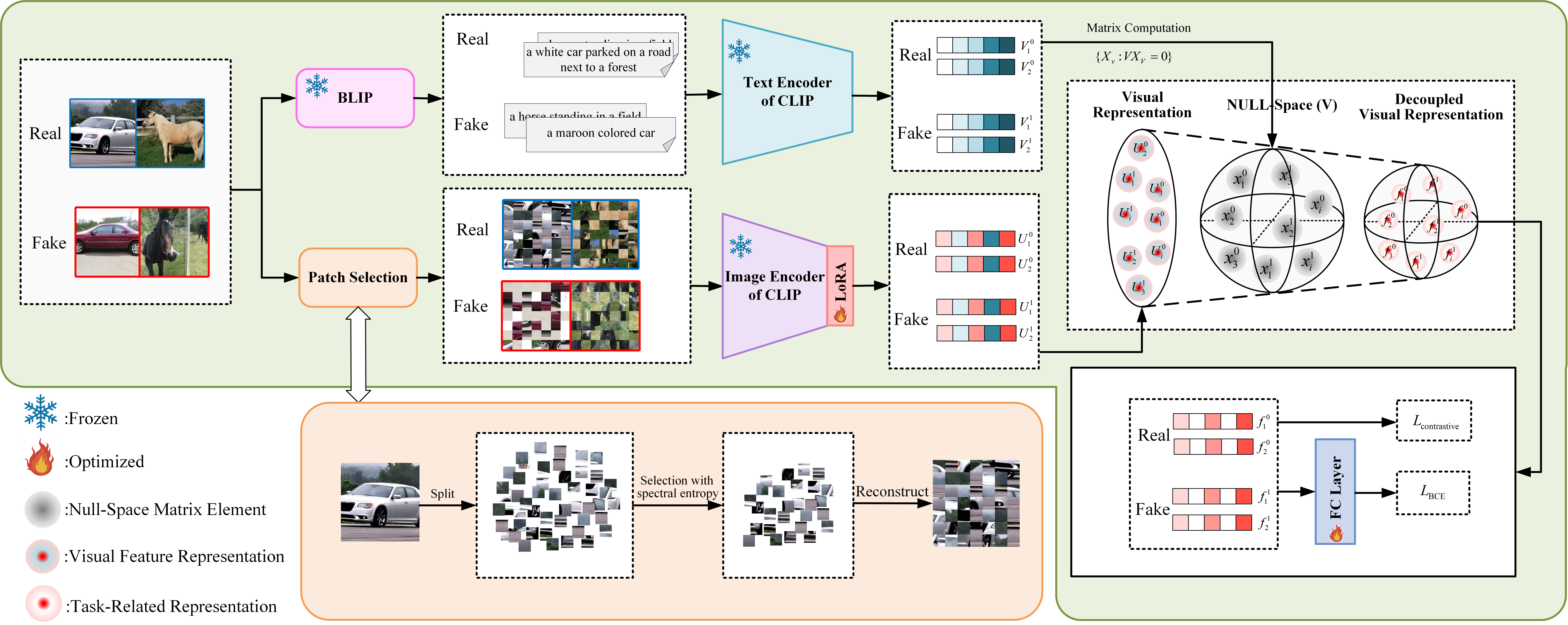}
    \caption{\textbf{Architecture of NS-Net for Generalizable AI-Generated Image Detection}. Specifically, we first employ the Patch Selection strategy adjusted for CLIP’s input size to preserve potential forgery-related artifacts. Subsequently, the visual features extracted by the CLIP's image encoder are projected onto the NULL-Space of the semantic information, effectively removing task-irrelevant semantic components. The resulting features, tailored for the detection task, are then utilized in a contrastive learning framework, which can not only guide the linear classification layer but also capture the intrinsic distributional differences between real and AI-generated images, enhancing the model’s ability to generalize beyond simple classification.}
    \label{fig:backbone}
\end{figure*}

\section{Methodology}
We propose the \textbf{NULL-Space Network (NS-Net)}, a novel framework designed to decouple semantic information embedded in CLIP features, preserving only the components relevant to forgery detection, which significantly improves the model’s generalization ability. As illustrated in Figure \ref{fig:backbone}, our architecture begins with a Patch Selection strategy to preserve fine-grained artifacts by selecting patches with the highest and lowest spectral entropy. We then project the visual features extracted from CLIP’s image encoder onto the NULL-Space of the semantic information, thereby effectively removing irrelevant or confounding semantics. The resulting task-specific features are subsequently leveraged in a contrastive learning setup, which not only guides the linear classification layer but also models the intrinsic distributional differences between real and AI-generated images, beyond simple categorical prediction. Furthermore, during inference, BLIP generates a text description for each input image, from which we compute the corresponding NULL-Space projection for semantic decoupling. The following sections elaborate on the proposed methodology in detail.

\subsection{Patch Selection}
\label{sec:patchselection}
In practice, input images are often extremely high-resolution. Although Vision Transformers (ViTs) are theoretically size-agnostic, their quadratic complexity with respect to sequence length makes large-scale processing computationally prohibitive; convolutional networks face similar limitations. Many studies adopt Resize to downsample inputs to an appropriate resolution, but this process inevitably suppresses artifact information\cite{li2025improving}. While CenterCrop \cite{li2025improving,tan2025c2p,yan2024orthogonal} is commonly used to alleviate the information loss caused by resizing, it only retains the artifacts at the center, overlooking critical forgery traces that may occur away from the center of the images. Different generation mechanisms exhibit complementary spectral-entropy characteristics: GANs often introduce high-frequency periodic peaks and sharpening residues, whereas diffusion models produce low-frequency distortions such as over-smoothing. To address these issues, we propose the Patch Selection strategy that maximizes the preservation of complementary high/low-entropy artifacts while adapting images to the fixed input size.

Specifically, we first divide the image into multiple non-overlapping patches using a fixed window size, denoted as $ I = \{x_1, x_2, ..., x_n\}, x_i \in {\mathbb{R}}^{N \times N \times 3} $, where $n$ is the total number of patches. In this work, we set the patch size to $ N = 32 $. Each patch is transformed into the frequency domain via Fast Fourier Transform (FFT), yielding $ I_f = \{x_1^f, x_2^f, ..., x_n^f\}, x_i^f \in {\mathbb{R}}^{N \times N \times 3} $. We then compute the spectral entropy of each patch to quantify its texture complexity and sort all patches in descending order.
Given the network’s required input image size $M \times M$, we determine the number of selected patches as $ 2K = (\frac{M}{N})^2 $, where $n \ge 2K$. We then select the top $K$ patches with the highest spectral entropy and the bottom $K$ patches with the lowest spectral entropy from the original $n$ patches. These patches are randomly shuffled and reconstructed into a new image $X_\text{{reconstruct}} \in {\mathbb{R}}^{M \times M \times 3} $ to match the input requirements. The proposed patch selection strategy not only ensures compatibility with networks requiring fixed-size inputs, but also preserves both the texture-rich and texture-poor regions, thereby reducing the loss of critical artifact information. Moreover, the shuffling process disrupts the original spatial semantics of the image, helping the model focus on subtle artifacts rather than high-level semantic content, thus enhancing its ability to detect forgeries. Different from existing PatchShuffle preprocessing methods \cite{zhong2023patchcraft,zheng2024breaking,gye2025reducing}, which still use Crop or Resize before shuffle, we adopt a "selective alternative cropping" strategy that directly selects patches rich in artifact information, thereby reducing information loss at the source and further enhancing the model’s generalization ability.

\subsection{NULL-Space Decoupling}
Preliminarily, the NULL-Space of a given matrix $A \in \mathbb{R} ^{B \times d}$, denoted as $\text{NULL-Space}(A)$ , is defined as:
\begin{equation}
    \text{NULL-Space}(A) = \{ X : AX = \mathbf{0}\}. 
\end{equation}
We also define the projection matrix $P$ as the operator that projects any feature vector $f$ onto the NULL-space of $A$, i.e., 
\begin{equation}
\forall f \in \mathbb{R} ^d, f'=Pf \implies f' \in X, Af'=\mathbf{0}.
\end{equation}
In particular, if $f$ belongs to the subspace of $A$, then the projection satisfies $Pf = \mathbf{0}$.

Analogously, our primary objective is to construct the NULL-Space corresponding to the semantic information embedded within the visual features extracted by CLIP’s image encoder. Specifically, we use the CLIP's image encoder to extract visual features $U$ from each image, which inherently encode both semantic information $U_s$ and artifact-related information $U_t$, as discussed in \cref{sec:Introduction}. However, the entanglement and complex interactions between $U_s$ and $U_t$ hinder the direct construction of the NULL-Space $X_{U_s}$ associated with semantic content alone. 
Since CLIP is pre-trained with an image–text alignment objective, matched image–text pairs are encouraged to have high similarity in the embedding space, implying that the semantic component of the visual feature is directionally consistent with the corresponding text embedding; thus, the text embedding $V$ can approximate the semantic component $U_s$, \textit{i.e.}, $V \approx U_s $. Consequently, we shift our goal to constructing the  NULL-Space $X_V$ of the textual features $V$, and then projecting the visual features $U$ onto $X_V$ in order to suppress semantic content and achieve effective disentanglement.
During training, we freeze both CLIP's image and text encoders and introduce LoRA (Low-Rank Adaptation) layers to fine-tune the image encoder. This design improves the encoder’s ability to emphasize artifact-related cues relevant for distinguishing AI-generated images.

\begin{table*}[!ht]
    \centering
    \large
    \caption{\textbf{Cross-model Accuracy (Acc.) Performance on the AIGCDetectBench \cite{zhong2023patchcraft} Dataset.} The first column represents the accuracy of detecting real images (R.Acc.), and the others are the accuracy of detecting fake images (F.Acc.).}
    \label{tab:AIGCDetectBench_Acc}
    \resizebox{1.0\linewidth}{!}{
    \begin{tabular}{lcccccccccccccccccccc}
    \bottomrule
        \multirow{3}*{Method}  & \multirow{3}*{\makecell[c]{Real\\Image}}    & \multicolumn{7}{c}{Generative Adversarial Networks}   & \multicolumn{2}{c}{Other} & \multicolumn{8}{c}{Diffusion Models}  &  \multirow{3}*{mAcc.}\\ 
        \cmidrule(r){3-9} \cmidrule(r){10-11} \cmidrule(r){12-19}
        ~                       & ~                     & \makecell[c]{Pro-\\GAN} & \makecell[c]{Cycle-\\GAN} & \makecell[c]{Big-\\GAN} & \makecell[c]{Style-\\GAN}  & \makecell[c]{Style-\\GAN2}  & \makecell[c]{Gau-\\GAN}  & \makecell[c]{Star-\\GAN}  & \makecell[c]{WFIR\\}  & \makecell[c]{Deep-\\fake} & {\makecell[c]{SDv1.4\\}}    & {\makecell[c]{SDv1.5\\}}    & {\makecell[c]{ADM}}    & {\makecell[c]{GLIDE}}   & {\makecell[c]{Mid-\\journey}}    & \makecell[c]{Wukong}     & \makecell[c]{VQDM}     & \makecell[c]{DALLE2}     & ~\\ \midrule
        CNN-Spot \cite{wang2020cnn} & 99.0          & 95.3      & 18.7      & 1.8       & 36.5      & 22.0      & 2.5       & 23.1      & 1.1       & 29.3      & 55.9      & 55.6      & 1.8       & 4.8       & 5.2       & 27.6      & 0.7       & 4.5       & 29.0      \\
        UnivFD \cite{ojha2023towards}& 92.3          & 98.9      & 90.5      & 79.2      & 55.7      & 48.7      & 91.1      & 96.9      & 92.8      & 26.9      & 96.3      & 96.0      & 12.7      & 75.6      & 61.2      & 84.7      & 45.6      & 62.3      & 72.7      \\
        FreqNet \cite{tan2024frequency}& 89.9          & 99.4      & 57.1      & 51.0      & 75.1      & 67.5      & 9.9       & 88.4      & 95.4      & 35.8      & 99.9      & 99.8      & 37.7      & 78.9      & 80.8      & 98.0      & 34.1      & 88.8      & 71.7      \\ 
        NPR \cite{tan2024rethinking}& 99.3          & 98.9      & 29.3      & 16.5      & 67.1      & 58.7      & 1.7       & 6.5       & 7.9       & 0.1       & 100.0     & 99.9      & 26.5      & 69.2      & 71.0      & 97.7      & 15.4      & 89.8      & 53.1      \\
        Ladeda \cite{cavia2024real} & 99.8          & 99.7      & 7.2       & 29.0      & 95.6      & 98.3      & 4.9       & 0.0       & 19.2      & 0.1       & 100.0     & 99.9      & 27.3      & 79.7      & \underline{88.8}      & 97.9      & 15.5      & 92.4      & 58.6      \\
        AIDE  \cite{yan2024sanity}  & 93.0          & 95.3      & 88.0      & 96.9      & 89.6      & 97.0      & 90.0      & 97.0      & 42.9      & 9.2       & 99.8      & 99.7      & \pmb{92.4}      & \pmb{98.7}      & 63.7      & 98.9      & 90.1      & \pmb{98.9}      & 85.6      \\
        C2P-CLIP* \cite{tan2025c2p} & \underline{99.8}          & 100.0     & 99.8      & 99.8      & 72.2      & 78.0      & 87.7      & 100.0     & 34.9      & 67.6      & 100.0     & 99.9      & 50.5      & 73.3      & 28.4      & 99.7      & 93.3      & 98.1      & 82.4      \\
        DFFreq \cite{yan2026dual}& 98.0          & 98.0      & 70.9      & 93.6      & \pmb{99.0}      & \pmb{98.9}      & 93.8      & 97.4      & 3.5       & 76.0      & 99.9      & 99.8      & 65.9      & 87.8      & \pmb{94.8}      & 99.1      & 76.0      & 95.9      & 85.2      \\
        SAFE \cite{li2025improving} & 99.2          & 99.7      & 85.6      & 83.5      & 88.1      & 96.7      & 89.0      & 99.9      & 8.4       & 17.3      & 99.8      & 99.6      & 36.8      & 90.5      & 86.3      & 98.5      & 84.0      & 92.0      & 80.8      \\
        VIB-Net \cite{zhang2025towards}& 93.5          & 99.4      & 97.3      & 91.3      & 71.3      & 73.3      & 97.8      & 97.3      & \pmb{97.3}      & \underline{90.6}      & 100.0     & 99.8      & 52.8      & 69.3      & 63.7      & 99.3      & 80.9      & 58.5      & 84.7      \\
        Effort \cite{yan2024orthogonal} & \pmb{99.8}          & \underline{100.0}     & \pmb{100.0}     & \underline{100.0}     & 79.6      & 82.1      & \pmb{100.0}      & \underline{100.0}     & 91.1     & 76.6      & \underline{100.0}     & \underline{99.9}      & 53.1      & 81.2      & 44.2      & 81.2      & \underline{96.3}      & 75.8      & 87.8      \\
        LTD \cite{yang2026layer}    & 98.7          & 99.9      & 99.8      & 99.7      & 76.1      & 76.0      & 96.9      & 100.0     & \underline{97.0}      & \pmb{94.3}      & 100.0     & 99.9      & 65.3      & 91.6      & 56.2      & \underline{99.8}      & 92.6      & 93.8      & \underline{91.0}      \\
        DoU \cite{he2026diversity}  & 98.3          & 94.1      & 97.5      & 99.1      & 67.3      & 37.9      & 82.1      & 100.0     & 77.9      & 78.7      & 100.0     & 99.9      & 66.7      & 84.7      & 45.9      & 99.8      & 95.6      & 85.2      & 83.9      \\ \midrule
        \rowcolor{light-gray} Ours  & 97.7          & \pmb{100.0}     & \underline{99.9}      & \pmb{100.0}     & \underline{95.7}      & \underline{98.6}      & \underline{98.5}      & \pmb{100.0}     & 68.9      & 60.6      & \pmb{100.0}     & \pmb{99.9}      & \underline{85.3}      & \underline{97.2}      & 77.4      & \pmb{100.0}     & \pmb{98.9}      & \underline{98.8}      & \pmb{93.2}      \\
    \bottomrule
    \end{tabular}
    }
\end{table*}

\begin{table*}[!ht]
    \centering
    \large
    \caption{\textbf{Cross-model Average Precision (A.P.) Performance on the AIGCDetectBench \cite{zhong2023patchcraft} Dataset.} }
    \label{tab:AIGCDetectBench_AP}
    \resizebox{1.0\linewidth}{!}{
    \begin{tabular}{lcccccccccccccccccc}
    \bottomrule
        \multirow{3}*{Method}  & \multicolumn{7}{c}{Generative Adversarial Networks}   & \multicolumn{2}{c}{Other} & \multicolumn{8}{c}{Diffusion Models}  &  \multirow{3}*{mA.P.}\\ 
        \cmidrule(r){2-8} \cmidrule(r){9-10} \cmidrule(r){11-18}
        ~                      & \makecell[c]{Pro-\\GAN} & \makecell[c]{Cycle-\\GAN} & \makecell[c]{Big-\\GAN} & \makecell[c]{Style-\\GAN}  & \makecell[c]{Style-\\GAN2}  & \makecell[c]{Gau-\\GAN}  & \makecell[c]{Star-\\GAN} & \makecell[c]{WFIR\\}  & \makecell[c]{Deep-\\fake} & {\makecell[c]{SDv1.4\\}}    & {\makecell[c]{SDv1.5\\}}    & {\makecell[c]{ADM}}    & {\makecell[c]{GLIDE}}   & {\makecell[c]{Mid-\\journey}}    & \makecell[c]{Wukong}     & \makecell[c]{VQDM}     & \makecell[c]{DALLE2}     & ~\\ \midrule
        CNN-Spot \cite{wang2020cnn} & 99.9      & 91.0      & 59.2      & 94.8      & 94.2      & 86.1      & 82.5      & 65.5      & 74.2      & 97.6      & 97.6      & 66.6      & 83.1      & 80.4      & 88.9      & 57.0      & 87.8      & 84.3      \\
        UnivFD \cite{ojha2023towards}& 99.9      & 90.5      & 79.2      & 55.7      & 48.7      & 91.1      & 96.9      & 92.8      & 26.9      & 96.3      & 96.0      & 65.2      & 95.3      & 91.7      & 97.4      & 87.2      & 62.3      & 91.9      \\
        FreqNet \cite{tan2024frequency}& 99.4      & 97.5      & 87.9      & 95.9      & 94.2      & 54.0      & 100.0     & 57.0      & 88.5      & 99.9      & 99.9      & 73.9      & 94.1      & 94.4      & 99.1      & 74.9      & 92.3      & 88.4      \\ 
        NPR  \cite{tan2024rethinking}& 100.0     & 98.6      & 79.4      & 98.6      & 99.6      & 71.3      & 97.6      & 65.5      & 92.5      & 100.0     & 99.9      & 74.1      & 97.4      & 97.8      & 99.9      & 81.5      & 99.3      & 91.4      \\
        Ladeda \cite{cavia2024real} & 100.0     & 91.5      & 93.2      & \underline{100.0}     & 100.0     & 81.8      & 93.0      & 86.9      & 72.8      & 100.0     & 99.9      & 77.3      & 99.1      & 99.7      & 100.0     & 89.3      & 99.8      & 93.2      \\
        AIDE \cite{yan2024sanity}   & 99.6      & 98.2      & 91.6      & 99.4      & 99.9      & 74.1      & 99.7      & 90.8      & 66.5      & 100.0     & 99.9      & \underline{99.4}      & \underline{99.9}      & 90.3      & 99.9      & 99.3      & 99.9      & 94.5      \\
        C2P-CLIP* \cite{tan2025c2p} & 100.0     & 100.0     & 99.9      & 99.3      & 99.7      & 98.8      & 100.0     & 99.6      & \pmb{98.3}      & 100.0     & 100.0     & 96.3      & 98.9      & 90.4      & 100.0     & 99.9      & 100.0     & 98.8      \\
        DFFreq \cite{yan2026dual}& 99.7      & 99.0      & 98.4      & \pmb{100.0}     & \underline{100.0}     & 98.1      & 99.8      & 89.0      & 87.5      & 100.0     & 99.9      & 98.7      & 99.5      & \underline{99.7}      & 100.0     & 99.3      & 99.9      & 98.1      \\
        SAFE \cite{li2025improving} & 100.0     & 99.5      & 97.8      & 99.9      & \pmb{100.0}     & 96.6      & 100.0     & 73.8      & 91.0      & 100.0     & 100.0     & 98.1      & \pmb{99.9}      & \pmb{99.9}      & 100.0     & 99.9      & \underline{100.0}     & 97.4      \\
        VIB-Net \cite{zhang2025towards}& 99.9      & 99.7      & 97.1      & 93.9      & 98.1      & 99.2      & 99.7      & 95.7      & 83.6      & 100.0     & 99.9      & 91.2      & 96.4      & 94.3      & 99.9      & 97.4      & 94.6      & 96.5      \\
        Effort \cite{yan2024orthogonal}& \underline{100.0}     & \underline{100.0}     & \pmb{100.0}     & 99.3      & 99.2      & \pmb{100.0}     & \underline{100.0}     & \pmb{100.0}     & 97.9      & \underline{100.0}     & \underline{100.0}     & 97.5      & 99.8      & 98.9      & \underline{100.0}     & \underline{100.0}     & 99.9      & \underline{99.7}      \\
        LTD \cite{yang2026layer}    & 100.0     & 100.0     & \underline{99.9}      & 99.8      & 99.8      & \underline{99.5}      & 100.0     & \underline{99.9}      & \underline{98.2}      & 100.0     & 100.0     & 99.2      & 99.8      & 98.7      & 100.0     & 99.9      & 99.9      & \pmb{99.7}      \\
        DoU \cite{he2026diversity}  & 100.0     & 99.9      & 99.7      & 98.0      & 96.3      & 95.1      & 100.0     & 99.9      & 91.7      & 100.0     & 100.0     & 99.4      & 99.9      & 99.2      & 100.0     & 100.0     & 100.0     & 98.8      \\ \midrule
        \rowcolor{light-gray} Ours  & \pmb{100.0}     & \pmb{100.0}     & 99.7      & 99.8      & 99.9      & 97.0      & \pmb{100.0}     & 99.5      & 95.2      & \pmb{100.0}     & \pmb{100.0}     & \pmb{100.0}     & 99.8      & 97.0      & \pmb{100.0}     & \pmb{100.0}     & \pmb{100.0}     & 99.2      \\
        \bottomrule
        \end{tabular}
    }
\end{table*}

To begin, we consider a dataset $\mathcal{D}$ consisting of both real and fake images, defined as:
\begin{equation}
    \mathcal{D} = \{\text{img}_j, \text{label}_j\}_{j=1}^N, \;\; \text{label} \in \{0, 1\},
\end{equation}
where $ \text{label} = 1 $ denotes a fake image, and $ \text{label} = 0 $ denotes a real image. For each image $ \text{img}_j $, we employ BLIP \cite{li2022blip} to generate a corresponding text description $ \text{text}_j $. This yields an enriched dataset with paired visual and textual information, defined as:
\begin{equation}
    \tilde{\mathcal{D}}  = \{\text{img}_j, \text{text}_j, \text{label}_j\}_{j=1}^N, \;\; \text{label} \in \{0, 1\}.
\end{equation}

Next, we extract visual features $U$ using CLIP’s image encoder and semantic features $V$ using the text encoder:
\begin{equation}
    U \;=\; \text{E}_{\text{I}}^{\text{LoRA}}(\text{img}), \;\;  V \;=\; \text{E}_{\text{T}}(\text{text}),
\end{equation}
where $\text{E}_{\text{I}}^{\text{LoRA}}$ denotes the image encoder augmented with LoRA layers, and $\text{E}_{\text{T}}$ represents the original text encoder. Ideally, semantic information $U_s$ in $U$ could be explicitly represented and removed via a NULL-Space projection. In practice, semantics and artifact cues are highly entangled in CLIP features, so directly estimating a clean semantic subspace from $U$ is non-trivial. Since $V$ serves as a semantic approximation of $U_s$, we substitute $V$ in place of $U_s$ and compute the NULL-Space $X_V$ of the textual features $V$:
\begin{equation}
    \text{NULL-Space}(V) \;=\; \{ X_V : VX_V = \mathbf{0}\}.
\end{equation}
In practice, we form $V$ batch-wise from text embeddings and use its NULL-Space projector to suppress semantic-aligned components in the corresponding visual features.

Specifically, we first apply Singular Value Decomposition (SVD) \cite{wang2021training} to $V$:
\begin{equation}
    \{M, \Sigma, N^{\mathsf{T}}\} \;=\; \text{SVD}\{V\},
\end{equation}
where $M$ and $N$ are orthogonal matrices containing the left and right singular vectors, respectively, and $\Sigma$ is a diagonal matrix with singular values on the diagonal. We remove the columns of $N$ corresponding to non-zero singular values in $\Sigma$, and denote the remaining submatrix as $\tilde{N}$, which spans the NULL-Space of $V$. Based on this, we define the projection matrix $P_V$ as:
\begin{equation}
    P_V \;=\; \tilde{N}(\tilde{N})^{\mathsf{T}}.
\end{equation}
The NULL-Space of $V$ is then given by:
\begin{equation}
\{ X_V : X_V = P_V \cdot v, \, v \in \mathbb{R}^d \},
\end{equation}
indicating that the projection matrix $P_V$ maps any input vector $v$ into the NULL-Space $X_V$. Notably, projecting the original text features $V$ yields zero, \textit{i.e.}, $ V P_V \;=\; \mathbf{0}.$

We then project the visual features $U$ onto the NULL-Space $X_V$ of the text features, calculated as $U P_V$. which effectively suppresses the semantic content aligned with the text modality, \textit{i.e.}, $ U_s P_V \approx \mathbf{0}$. Importantly, linear operations such as projection onto a NULL-Space can transform the feature space while preserving orthogonal (i.e., non-semantic) information. Therefore, $U P_V$ achieves semantic disentanglement by filtering out semantic components, while retaining information that is potentially discriminative for forgery detection.

\subsection{Contrastive Learning and Classification}
In order to make the detector more focused on learning the difference between the features of real and generated images, we use contrastive loss \cite{chen2020simple} $\mathcal{L}_{\text{contrastive}}$ as the dominant and linear classifier loss $\mathcal{L}_{\text{BCE}}$ as the auxiliary for training.

\begin{table*}[ht!]
    \centering
    \large
    \caption{\textbf{Cross-Diffusion-Sources Evaluation on the Diffusion Test Set of UniversalFakeDetect \cite{ojha2023towards}.} }
    \label{tab:UniversalFakeDetect}
    \resizebox{1.0\linewidth}{!}{
        \begin{tabular}{lcccccccccccccccc|cccc}
        \bottomrule
        \multirow{2}{*}{Method} & \multicolumn{2}{c}{Guided} & \multicolumn{2}{c}{Glide\_50\_27} & \multicolumn{2}{c}{Glide\_100\_10} & \multicolumn{2}{c}{Glide\_100\_27} & \multicolumn{2}{c}{LDM\_100} & \multicolumn{2}{c}{LDM\_200} & \multicolumn{2}{c}{LDM\_200\_cfg} & \multicolumn{2}{c}{DALLE} & \multicolumn{4}{|c}{Mean} \\ \cline{2-21} 
                                & F.Acc.       & A.P.        & F.Acc.       & A.P.         & F.Acc.    & A.P.         & F.Acc.       & A.P.          & F.Acc.    & A.P.        & F.Acc.        & A.P.        & F.Acc.       & A.P.         & F.Acc.    & A.P.      & R.Acc.       & F.Acc.    & Acc.       & A.P.               \\ \midrule
        CNN-Spot \cite{wang2020cnn} & 3.0          & 70.8        & 4.6          & 76.9         & 4.6       & 78.0         & 5.1          & 76.8          & 21.4      & 88.7        & 24.1          & 90.1        & 36.1         & 93.8         & 3.5       & 62.7      & 99.8         & 12.8      & 56.3       & 79.7               \\
        UnivFD \cite{ojha2023towards}& 35.5         & 81.0        & 79.1         & 92.5         & 80.0      & 92.1         & 79.6         & 92.3          & 75.9      & 91.4        & 76.0          & 91.4        & 57.7         & 84.7         & 49.1      & 80.3      & 89.8         & 66.4      & 78.1       & 88.2               \\
        FreqNet \cite{tan2024frequency}& 40.9         & 78.5        & 80.4         & 97.8         & 81.1      & 97.9         & 78.2         & 97.6          & 96.6      & 99.7        & 97.6          & 99.8        & 96.1         & 99.7         & 40.1      & 93.1      & 97.9         & 76.4      & 87.6       & 95.5               \\
        NPR \cite{tan2024rethinking}& 29.5         & 83.0        & 61.9         & 99.3         & 60.6      & 99.0         & 59.9         & 99.3          & 76.1      & 99.9        & 73.9          & 99.9        & 89.3         & 100.0        & 27.3      & 98.9      & 99.9         & 59.8      & 79.8       & 97.4               \\
        Ladeda \cite{cavia2024real}& 39.5         & 89.0        & 69.4         & 99.6         & 71.6      & 99.4         & 72.4         & 99.7          & 74.3      & 99.9        & 75.7          & 99.9        & 85.3         & 99.9         & 9.2       & 96.4      & 100.0        & 62.2      & 81.1       & 98.0               \\
        AIDE  \cite{yan2024sanity}& \underline{81.7}         & 98.5        & \pmb{97.3}         & 99.9         & \underline{96.5}      & 99.8         & \pmb{97.0}         & 99.8          & 98.6      & 99.9        & 98.4          & 100.0       & 99.7         & 100.0        & 96.1      & 99.5      & 97.4         & \underline{95.6}      & 96.5       & 99.6               \\
        C2P-CLIP* \cite{tan2025c2p}& 60.7         & 98.6        & 88.7         & 99.6         & 86.0      & 99.8         & 78.6         & 99.4          & \underline{100.0}     & 100.0       & 100.0         & 100.0       & 100.0        & 100.0        & 100.0     & 100.0     & 100.0         & 89.2      & 94.6      & 99.7               \\
        DFFreq \cite{yan2026dual}& 80.7         & 99.2        & 90.7         & 99.2         & 92.5      & 99.2         & 88.6         & 99.0          & 99.9      & 100.0       & 99.8          & 100.0       & 99.7         & 100.0        & 91.4      & 99.5      & 98.8         & 92.9      & 96.0       & 99.5               \\
        SAFE \cite{li2025improving}& 45.5         & 98.9        & 92.9         & 99.6         & 94.7      & 99.7         & 90.2         & 99.4          & 99.8      & 100.0       & 99.9          & 100.0       & 99.7         & 100.0        & 96.1      & 99.9      & 99.3         & 89.9      & 94.6       & 99.7               \\
        VIB-Net \cite{zhang2025towards}& 58.1         & 94.2        & 75.5         & 97.2         & 76.6      & 96.8         & 72.4         & 96.9          & 97.1      & 99.9        & 97.0          & 99.8        & 86.5         & 98.6         & 93.9      & 99.5      & 98.8         & 82.1      & 90.5       & 97.9               \\ 
        Effort \cite{yan2024orthogonal} & 53.0         & 98.3        & 91.4         & 99.8         & 91.4      & 99.8         & 87.4         & 99.6          & 99.8      & 100.0       & \underline{100.0}         & 100.0       & 99.6         & 100.0        & \underline{100.0}     & 100.0     & \underline{100.0}        & 90.3      & 95.2       & 99.7               \\
        LTD \cite{yang2026layer} & 67.3         & \underline{99.5}        & 95.2         & \pmb{100.0}         & 94.2      & \pmb{100.0}         & 91.8         & \pmb{99.9}          & 99.3      & 100.0       & 99.3          & 100.0       & 98.3         & 100.0        & 99.9      & 100.0     & \pmb{100.0}        & 93.2      & \underline{96.6}       & 99.9               \\
        DoU \cite{he2026diversity}      & 58.6         & \pmb{99.5}        & 80.3         & 99.8         & 81.7      & 99.8         & 77.8         & 99.8          & 99.9      & \underline{100.0}       & 99.9          & \underline{100.0}       & \underline{100.0}        & \underline{100.0}        & 99.4      & \underline{100.0}     & 99.9         & 87.2      & 93.5       & \underline{99.9}               \\ \midrule
    \rowcolor{light-gray}Ours   & \pmb{85.3}         & 99.4        & \underline{96.7}         & \underline{99.9}         & \pmb{98.0}      & \underline{99.9}         & \underline{96.4}         & \underline{99.8}          & \pmb{100.0}     & \pmb{100.0}       & \pmb{100.0}         & \pmb{100.0}       & \pmb{100.0}        & \pmb{100.0}        & \pmb{100.0}     & \pmb{100.0}     & 99.9         & \pmb{97.0}      & \pmb{98.5}       & \pmb{99.9}               \\
        \bottomrule
        \end{tabular}
    }
\end{table*}

\begin{table*}[!ht]
    \centering
    \large
    \caption{\textbf{Cross-model Accuracy (Acc.) Performance on the AIGIBench \cite{li2025artificial} Dataset.} The first row represents the accuracy of detecting real images (R.Acc.), and the others are the accuracy of detecting fake images (F.Acc.).}
    \label{tab:AIGIBench_Acc}
    \resizebox{1.0\linewidth}{!}{
    \begin{tabular}{lccccccccc>{\columncolor{light-gray}}c}
    \bottomrule
        Generator           & CNN-Spot \cite{wang2020cnn}& UnivFD \cite{ojha2023towards}& FreqNet \cite{tan2024frequency}& NPR \cite{tan2024rethinking} & Ladeda \cite{cavia2024real} & AIDE \cite{yan2024sanity} & DFFreq \cite{yan2026dual} & SAFE \cite{li2025improving} & VIB-Net \cite{zhang2025towards} & Ours      \\ \midrule
        Real Image          & \pmb{98.0}            & 73.2       & 65.8       & 93.8      & \underline{97.1}          & 88.1      & 91.7      & 92.2      & 60.8          & 88.0      \\ \midrule
        R3GAN               & 2.3             & 94.1       & 59.9       & 8.4       & 19.5          & \pmb{99.0}      & 78.4      & 93.0      & 79.3          & \underline{98.5}      \\ 
        StyleGAN3           & 9.1             & 82.6       & \underline{98.2}       & 63.6      & 93.2          & 91.1      & 95.5      & 94.6      & 89.8          & \pmb{99.6}      \\
        StyleGAN-XL         & 0.7             & 96.7       & 95.5       & 28.2      & 80.5          & 91.7      & 15.6      & 89.4      & \pmb{99.8}          & \underline{99.3}      \\ 
        StyleSwim           & 6.9             & 98.1       & 97.1       & 77.7      & 97.3          & 82.0      & 99.8      & \pmb{99.9}      & 98.4          & \underline{99.8}      \\ 
        FLUX1-dev           & 16.3            & 86.6       & 92.4       & 97.2      & \underline{99.3}          & 90.0      & 96.1      & \pmb{99.5}      & 60.5          & 96.2      \\ 
        Midjourney-V6       & 5.8             & 80.6       & 83.6       & 53.8      & 83.4          & 79.8      & \pmb{95.8}      & 94.1      & 80.2          & \underline{94.8}      \\
        GLIDE               & 4.6             & 75.2       & 79.7       & 70.3      & 81.8          & \pmb{98.4}      & 86.9      & 89.2      & 69.2          & \underline{96.6}      \\
        Imagen3             & 4.2             & 84.2       & 81.5       & 78.2      & 92.6          & 93.9      & 51.9      & \underline{94.8}      & 82.5          & \pmb{99.5}      \\
        SD3                 & 13.3            & 90.6       & 88.1       & 89.7      & 99.0          & \underline{99.3}      & 92.1      & 94.4      & 82.1          & \pmb{99.9}      \\
        SDXL                & 7.2             & 88.0       & 98.9       & 79.0      & 98.3          & 97.6      & 98.0      & \underline{99.9}      & 89.3          & \pmb{99.9}      \\
        BLIP                & 56.5            & 92.1       & 100.0      & 99.9      & 100.0         & 100.0     & 100.0     & \underline{100.0}     & 99.6          & \pmb{100.0}     \\
        Infinite-ID         & 1.1             & 93.8       & 92.7       & 34.6      & 32.2          & \underline{97.5}      & 93.9      & \pmb{99.2}      & 72.4          & 92.2      \\
        IP-Adapter          & 6.0             & 92.0       & 92.0       & 71.8      & 90.6          & 93.5      & \underline{97.8}      & 97.2      & 93.1          & \pmb{99.8}      \\
        SocialRF            & 7.5             & \pmb{55.5}       & \underline{39.3}       & 21.9      & 19.4          & 18.4      & 18.4      & 17.1      & 27.4          & 18.4      \\ 
        CommunityAI         & 5.4             & \pmb{51.2}       & 12.2       & 8.2       & 9.0           & 9.3       & 9.2       & 9.1       & \underline{13.8}          & 8.9       \\ \midrule
        Average             & 15.3            & 83.4       & 79.8       & 61.0      & 74.5          & 83.1      & 76.3      & \underline{85.2}      & 74.9          & \pmb{87.0}      \\ \bottomrule
    \end{tabular}
    }
\end{table*}

Specifically, for the decoupled features, we define features with the same label as positive samples and features with different labels as negative. We hope that positive samples are closer, while negative samples are gradually farther away. Thus, let $i \in I \equiv \{1...N\}$ be the index of the batch sample, where N is the batchsize. The self-supervised contrastive loss of decoupled features can be formulated as follows:
\begin{equation}
    \mathcal{L}_{\text{contrastive}} = -\frac{1}{|\mathcal{P}_i|} \sum_{p \in \mathcal{P}_i} \log \left( \frac{\exp(f_{i} \cdot f_{p}/ \tau)}{\sum_{j=1, j \neq i}^N \exp(f_{i} \cdot f_{j} / \tau)} \right),
\end{equation}
where $f$ are the decoupled features, the index $i$ called the anchor, the index $p$ called the positive, $ \mathcal{P}_i = \{ j \mid \text{label}_i = \text{label}_j, \, j \neq i \} $ is the set of positive samples, $\tau$ is a temperature hyperparameter and $|\cdot|$ denotes the number of vectors. 

The final loss function is obtained by the weighted sum of the above loss functions as follows:
\begin{equation}
    \mathcal L \;=\; (1 - \lambda) * \mathcal L_{\text{contrastive}} + \lambda * \mathcal L_{\text{BCE}}, 
\end{equation}
where $\lambda$ is the hyper-parameter for balancing two losses.

\section{Experiments}
\subsection{Evaluation Setup}

\noindent\textbf{Datasets.} 
We utilize the dataset from AIGIBench \cite{li2025artificial} for training, specifically adopting \textbf{Setting-II}, which involves training on 144k images generated by ProGAN \cite{karras2018progressive} and SDv1.4 \cite{rombach2022high}. In addition, to comprehensively evaluate the effectiveness of our method, we conduct generalization experiments on four different datasets, including AIGCDetectBench \cite{zhong2023patchcraft}, UniversalFakeDetect \cite{ojha2023towards}, AIGIBench \cite{li2025artificial}, and ARForensics \cite{zhang2025d}.

\noindent\textbf{Evaluation Metrics.} Following the established evaluation paradigm \cite{li2025artificial}, we adopt the accuracy of detecting fake images (F.Acc.) and average precision (A.P.) as primary evaluation metrics. Additionally, we calculate the accuracy of detecting real images (R.Acc.) at the beginning of the table to provide a complete assessment of the model's performance.

\noindent\textbf{Implementation Details.} We use the Adam optimizer with a learning rate of $1 \times 10^{-4}$. We train the model for only 2 epochs with the batchsize of 32. The ViT-L/14 model of CLIP is adopted as the pre-trained model. We add Low-Rank Adaptation to fine-tune CLIP's image encoder \cite{zanella2024low} with the settings of the hyperparameters as follows: \text{lora\_r = 6}, \text{lora\_alpha = 6}, and \text{lora\_dropout = 0.8}. The hyperparameter $\lambda$ is set to 0.2. In addition, to ensure a fair comparison, we retrain all methods on the AIGIBench training set.

\begin{table*}[!ht]
    \centering
    \large
    \caption{\textbf{Cross-source Accuracy (Acc.) and Average Precision (A.P.) Performance on the ARForensics Dataset.}}
    \label{tab:ARForensics}
    \resizebox{1.0\linewidth}{!}{
    \begin{tabular}{lcccccccccccccc|cc}
    \bottomrule
        \multirow{2}{*}{Method}
        & \multicolumn{2}{c}{LlamaGen}
        & \multicolumn{2}{c}{Janus-Pro}
        & \multicolumn{2}{c}{VAR}
        & \multicolumn{2}{c}{Infinity}
        & \multicolumn{2}{c}{Open-MAGVIT2}
        & \multicolumn{2}{c}{Switti}
        & \multicolumn{2}{c}{RAR}
        & \multicolumn{2}{|c}{Mean} \\ \cline{2-17}
        & Acc. & A.P. & Acc. & A.P. & Acc. & A.P. & Acc. & A.P. & Acc. & A.P. & Acc. & A.P. & Acc. & A.P. & Acc. & A.P. \\ \midrule
        CNN-Spot \cite{wang2020cnn}
        & 50.9 & 72.4 & 63.3 & 91.6 & 50.1 & 61.5 & 63.4 & 94.5 & 50.2 & 58.2 & 54.3 & 81.9 & 50.6 & 68.7 & 54.7 & 75.5 \\
        UnivFD \cite{ojha2023towards}
        & 71.2 & 85.0 & 80.7 & 91.3 & 61.1 & 74.9 & 91.8 & 97.0 & 59.3 & 72.7 & 91.2 & 96.9 & 83.2 & 93.2 & 76.9 & 87.3 \\
        FreqNet \cite{tan2024frequency}
        & 90.3 & 96.9 & 95.6 & 99.8 & 78.3 & 89.2 & 95.5 & 99.7 & 76.4 & 88.3 & 95.3 & 99.8 & 80.8 & 91.7 & 87.5 & 95.1 \\
        NPR \cite{tan2024rethinking}
        & 83.5 & 97.5 & 97.0 & 99.8 & 81.8 & 98.6 & 89.7 & 99.5 & 55.7 & 82.0 & 99.4 & 100.0 & 67.7 & 93.6 & 82.1 & 95.9 \\
        Ladeda \cite{cavia2024real}
        & 89.4 & 99.6 & \underline{99.6} & 100.0 & 80.5 & 98.7  & 99.8 & 100.0 & 78.0 & 99.0 & 99.8 & 100.0 & 87.1 & 99.6 & 90.6 & 99.5 \\
        AIDE \cite{yan2024sanity}
        & 91.0 & 97.8 & 96.3 & 99.4 & 92.8 & 98.4   & 95.3 & 98.8 & \pmb{93.4} & 98.6 & 94.3 & 99.0 & 92.5 & 98.4 & 93.7 & 98.6 \\
        DFFreq \cite{yan2026dual}
        & 96.2 & 99.9 & 99.2 & 100.0 & 91.1 & 99.8  & 96.9 & 99.9 & 78.4 & 99.0 & 99.5 & 100.0  & 96.6 & 100.0 & 94.0 & 99.8 \\
        SAFE \cite{li2025improving}
        & 96.4 & 100.0 & 99.8 & 100.0 & \pmb{99.1} & \pmb{100.0}    & 99.8 & 100.0 & \underline{91.4} & \pmb{99.9}  & \underline{99.9} & 100.0 & 96.7 & \pmb{100.0} & \underline{97.6} & \pmb{100.0} \\
        VIB-Net \cite{zhang2025towards}
        & 86.2 & 96.1 & 98.4 & 99.7 & 67.0 & 85.9   & 98.9 & 99.9 & 75.1 & 90.9 & 95.3 & 99.3   & 93.2 & 98.4 & 87.7 & 95.8 \\
        Effort \cite{yan2024orthogonal}
        & 97.5 & \underline{100.0} & \underline{100.0} & 100.0  & 92.5 & 99.8 & 99.6 & 100.0 & 81.8 & 97.8  & 98.4 & 100.0 & \underline{99.8} & 100.0   & 95.7 & 99.7 \\
        LTD \cite{yang2026layer}
        & 95.6 & 99.9 & 99.8 & 100.0 & 85.2 & 99.5  & 99.7 & 100.0 & 80.6 & 99.1 & 98.2 & 99.9  & 99.6 & 100.0 & 94.1 & 99.8 \\
        DoU \cite{he2026diversity}
        & \underline{98.3} & \pmb{100.0} & \pmb{100.0} & \underline{100.0}  & 94.4 & \underline{99.9} & \underline{99.8} & \underline{100.0}    & 84.0 & \underline{99.7} & 99.0 & \underline{100.0}    & 98.9 & \underline{100.0} & 96.3 & \underline{99.9} \\ \midrule
        \rowcolor{light-gray} Ours
        & \pmb{99.2} & 99.9 & 99.9 & \pmb{100.0}    & \underline{98.1} & 99.8 & \pmb{100.0} & \pmb{100.0}   & 91.2 & 98.0 & \pmb{99.9} & \pmb{100.0}    & \pmb{99.9} & \pmb{100.0} & \pmb{98.3} & 99.7 \\
    \bottomrule
    \end{tabular}
    }
\end{table*}

\begin{table}[!]
\centering
\large
\caption{\textbf{Ablation Study on the AIGCDetectBench \cite{zhong2023patchcraft} and AIGIBench \cite{li2025artificial} Dataset.} }
\label{tab:Ablation}
\resizebox{1.0\linewidth}{!}{
    \begin{tabular}{ccc|cc}
    \bottomrule
    \makecell[c]{Patch\\Selection}       & \makecell[c]{NULL-Space\\Decoupling}      & \makecell[c]{Contrastive\\Learning}       & \makecell[c]{AIGCDetect-\\Bench}      & AIGIBench     \\ \midrule
    \xmark               &\xmark                      &\xmark                      & 81.2                 & 71.4          \\
    \cmark               &\xmark                      &\xmark                      & 82.8                 & 74.7          \\
    \xmark               &\cmark                      &\xmark                      & 85.3                 & 79.8          \\
    \xmark               &\xmark                      &\cmark                      & 84.2                 & 77.8          \\
    \cmark               &\xmark                      &\cmark                      & \underline{87.5}                 & \underline{80.4}          \\ \midrule
    \rowcolor{light-gray}   \cmark      &\cmark       &\cmark                      & \textbf{93.2}                 & \textbf{87.0}          \\ \bottomrule
    \end{tabular}
}
\end{table}

\begin{table}[!]
\centering
\normalsize
\caption{\textbf{Selection Strategy Comparison on the AIGCDetectBench \cite{zhong2023patchcraft} and AIGIBench \cite{li2025artificial} Datasets.}}
\label{tab:selection_comparison}
\resizebox{1.0\linewidth}{!}{
    \begin{tabular}{c|cc}
    \bottomrule
    Selection Strategy                      & AIGCDetectBench      & AIGIBench \\ \midrule
    Resize                                  & 78.6          & 70.1      \\
    CenterCrop                              & 89.1          & 75.9      \\
    PatchShuffle                            & 87.3          & 76.9      \\
    PatchCraft                              & 77.1          & 60.3      \\ \midrule
    Random 2k patches                       & 88.4          & \underline{81.2}      \\
    Top 2k patches                          & 85.8          & 75.1      \\
    Bottom 2k patches                       & \underline{90.1}          & 78.8      \\ \midrule
    $14 \times 14$ patchsize                & 89.4          & 83.4      \\
    $16 \times 16$ patchsize                & 91.5          & 85.6      \\ \midrule
    \rowcolor{light-gray} Ours              & \textbf{93.2}          & \textbf{87.0}      \\ \bottomrule
    \end{tabular}
}
\end{table}

\subsection{Comparison with the State-of-the-Art}
\noindent\textbf{Performance on AIGCDetectBench.} As shown in Table \ref{tab:AIGCDetectBench_Acc} and Table \ref{tab:AIGCDetectBench_AP}, our method demonstrates superior cross-domain generalization on generated images, achieving 93.2\% mAcc. and 99.2\% mA.P. While its accuracy on real images (97.7\%) is slightly lower than some methods, it surpasses the state-of-the-art LTD \cite{yang2026layer} by 2.2\% in mean accuracy (mAcc.). This generalizability highlights the effectiveness of our approach in decoupling semantic information to focus on features critical for forgery detection.

\noindent\textbf{Performance on UniversalFakeDetect.} Table \ref{tab:UniversalFakeDetect} underscores our method's capability in detecting unknown Diffusion-based fake images, achieving an average accuracy of 98.5\%, surpassing the state-of-the-art LTD \cite{yang2026layer} by 1.9\% in mean accuracy (mAcc.). This consistent high performance across diverse diffusion models demonstrates the method's strong generalization ability.

\begin{table}[!]
    \centering
    \caption{\textbf{Robustness on JPEG Compression and Gaussian Blur of Our Method on the AIGCDetectBench \cite{zhong2023patchcraft} Dataset.}}
    \label{table:Robustness_AIGCDetectBench}
    \resizebox{1.0\linewidth}{!}{
    \begin{tabular}{l|ccc|ccc}
    \bottomrule
    \multirow{2}{*}{Method}         & \multicolumn{3}{c|}{JPEG Compression}  & \multicolumn{3}{c}{Gaussian Blur}                     \\
                                & QF=95         & QF=85         & QF=75             & $\sigma = 0.5$        & $\sigma = 1.0$        & $\sigma = 1.5$        \\ \midrule
    UnivFD                      & 76.4          & 74.2          & 73.8              & 80.2                  & 76.8                  & 76.3                  \\
    NPR                         & 72.8          & 70.7          & 69.2              & 80.6                  & 79.6                  & 81.0                  \\
    AIDE                        & 74.6          & 74.1          & 70.3              & \underline{88.6}                  & 75.8                  & 81.1                  \\
    SAFE                        & 57.7          & 59.6          & 60.4              & 88.5                  & 80.6                  & \underline{83.0}                  \\
    VIB-Net                     & \underline{82.2}          & \underline{80.8}          & \underline{79.0}              & 86.1                  & \underline{81.1}                  & 81.6                  \\
    \rowcolor{light-gray}Ours   & \textbf{85.6}          & \textbf{82.3}          & \textbf{79.6}              & \textbf{88.8}                  & \textbf{86.1}                  & \textbf{85.9}                  \\ \bottomrule
    \end{tabular}
    }
\end{table}

\begin{table}[!]
    \centering
    \caption{\textbf{Robustness on JPEG Compression and Gaussian Blur of Our Method on the AIGIBench \cite{li2025artificial} Dataset.}}
    \label{table:Robustness_AIGIBench}
    \resizebox{1.0\linewidth}{!}{
    \begin{tabular}{l|ccc|ccc}
    \bottomrule
    \multirow{2}{*}{Method}         & \multicolumn{3}{c|}{JPEG Compression}  & \multicolumn{3}{c}{Gaussian Blur}                     \\
                                & QF=95         & QF=85         & QF=75             & $\sigma = 0.5$        & $\sigma = 1.0$        & $\sigma = 1.5$        \\ \midrule
    UnivFD                      & \underline{65.6}          & \underline{66.3}          & \underline{65.9}              & 72.2                  & \underline{68.6}                  & 68.6                  \\
    NPR                         & 59.0          & 58.9          & 58.6              & 62.9                  & 62.0                  & 62.9                  \\
    AIDE                        & 61.0          & 59.6          & 55.9              & \underline{75.5}                  & 63.4                  & 65.2                  \\
    SAFE                        & 51.3          & 47.9          & 47.8              & \textbf{75.5}                  & 67.6                  & \underline{69.3}                  \\
    VIB-Net                     & 62.9          & 61.9          & 63.0              & 66.5                  & 62.6                  & 62.6                  \\ \midrule
    \rowcolor{light-gray}Ours   & \textbf{68.9}          & \textbf{67.5}          & \textbf{66.9}              & 72.8                  & \textbf{70.3}                  & \textbf{70.2}                  \\ \bottomrule
    \end{tabular}
    }
\end{table}

\noindent\textbf{Performance on AIGIBench.} Given the rapid evolution of generative techniques, existing detection datasets often fall short in evaluating contemporary AI-generated image detection methods. We address this by utilizing the AIGIBench dataset \cite{li2025artificial}, which includes a wide range of advanced generative models (Table \ref{tab:AIGIBench_Acc}). Our method still maintains the highest detection accuracy, with the improvement of 1.8\% compared with the state-of-the-art methods. It significantly outperforms UnivFD by over 10\% on most datasets, with a 21.4\% improvement on GLIDE, owing to its ability to eliminate semantic information from CLIP’s upstream training while retaining fine-grained features essential for detection.  

\noindent\textbf{Performance on ARForensics.}
Besides GANs and diffusion models, autoregressive image generators \cite{tian2024visual,tang2410hart} have recently emerged as an important class of generative models. To evaluate the generalization ability of NS-Net on autoregressive images, we conduct experiments on the ARForensics \cite{zhang2025d} dataset. As shown in Table~\ref{tab:ARForensics}, NS-Net achieves the highest mean accuracy of 98.3\%, outperforming the strongest compared baseline, SAFE, by 0.7\%.

\subsection{Ablation Studies}
To thoroughly assess the effectiveness of our method for AI-generated image detection, we conducted an ablation study of each module, which is illustrated in Table \ref{tab:Ablation}.

\noindent\textbf{NULL-Space Decoupling.} Adding only NULL-Space Decoupling to the baseline increases the mean accuracy to 85.3\% and 79.8\%, a 4.1\% and 8.4\% improvement on AIGCDetectBench and AIGIBench respectively. This enhancement, achieved without additional training, suggests that NULL-Space Decoupling effectively removes task-irrelevant semantic information from the feature space, allowing the model to focus on features more pertinent to distinguishing real and fake images. Its importance is further highlighted when removed from the full model, where the mean accuracy on AIGCDetectBench drops significantly from 93.2\% to 87.5\%, a substantial decrease of 5.7\%. This demonstrates the critical role it plays in maintaining high detection performance.

\begin{table*}[!ht]
    \centering
    \caption{\textbf{Plug \& Play Application of NULL-Space Decoupling on Existing Detectors on the AIGCDetectBench \cite{zhong2023patchcraft} Dataset.}}
    \label{tab:plug_decoupling}
    \resizebox{1.0\linewidth}{!}{
    \begin{tabular}{c|cccccccc|c}
    \bottomrule
    Method                              & SDv1.4    & SDv1.5    & ADM       & GLIDE     & Midjourney    & Wukong    & VQDM      & DALLE2    & mAcc. \\ \midrule
    UnivFD \cite{ojha2023towards}       & 96.3      & 96.0      & 12.7      & 75.6      & 61.2          & 84.7      & 45.6      & 62.3      & 66.8  \\
    \rowcolor{light-gray}+NULL-Space    & 97.8      & 97.4      & 59.5      & 74.9      & 65.9          & 95.4      & 81.4      & 73.0      & \pmb{$80.7_{+13.9}$}  \\ \midrule
    VIB-Net \cite{zhang2025towards}     & 100.0     & 99.8      & 52.8      & 69.3      & 63.7          & 99.3      & 80.9      & 58.5      & 78.0  \\ 
    \rowcolor{light-gray}+NULL-Space    & 100.0     & 99.8      & 55.1      & 72.3      & 68.0          & 99.5      & 83.3      & 63.3      & \pmb{$80.2_{+2.2}$}  \\ \midrule
    Ours w/o NULL-Space                 & 100.0     & 99.9      & 73.1      & 90.7      & 72.3          & 99.3      & 91.0      & 90.4      & 89.6  \\
    \rowcolor{light-gray}Ours           & 100.0     & 99.9      & 85.3      & 97.2      & 77.4          & 100.0     & 98.9      & 98.8      & \pmb{$94.7_{+5.1}$}  \\
    \bottomrule
    \end{tabular}
    }
\end{table*}

\begin{table*}[!ht]
    \centering
    \caption{\textbf{Plug \& Play Application of Patch Selection on Existing Detectors on the AIGCDetectBench \cite{zhong2023patchcraft} Dataset.}}
    \label{tab:plug_patch}
    \resizebox{1.0\linewidth}{!}{
    \begin{tabular}{c|cccccccc|c}
    \bottomrule
    Method                                  & SDv1.4    & SDv1.5    & ADM       & GLIDE     & Midjourney    & Wukong    & VQDM      & DALLE2    & mAcc. \\ \midrule
    DFFreq \cite{ojha2023towards}           & 99.9      & 99.8      & 65.9      & 87.8      & 94.8          & 99.1      & 76.0      & 95.9      & 89.9  \\
    \rowcolor{light-gray}+Patch Selection   & 99.9      & 99.9      & 69.1      & 93.2      & 89.5          & 99.6      & 84.2      & 96.3      & \pmb{$91.5_{+1.6}$}  \\ \midrule
    SAFE \cite{li2025improving}             & 99.8      & 99.6      & 36.8      & 90.5      & 86.3          & 98.5      & 84.0      & 92.0      & 85.9  \\
    \rowcolor{light-gray}+Patch Selection   & 100.0     & 99.8      & 54.1      & 95.3      & 95.8          & 99.4      & 84.4      & 97.7      & \pmb{$90.7_{+4.8}$}  \\ \midrule
    VIB-Net \cite{zhang2025towards}         & 100.0     & 99.8      & 52.8      & 69.3      & 63.7          & 99.3      & 80.9      & 58.5      & 78.0  \\ 
    \rowcolor{light-gray}+Patch Selection   & 100.0     & 99.9      & 73.9      & 85.3      & 66.8          & 99.0      & 83.9      & 79.6      & \pmb{$86.1_{+8.1}$}  \\ \midrule
    Ours w/o Patch Selection                & 100.0     & 99.8      & 74.7      & 78.8      & 54.7          & 99.9      & 94.8      & 92.4      & 86.9  \\
    \rowcolor{light-gray}Ours               & 100.0     & 99.9      & 85.3      & 97.2      & 77.4          & 100.0     & 98.9      & 98.8      & \pmb{$94.7_{+7.8}$}  \\
    \bottomrule
    \end{tabular}
    }
\end{table*}

\begin{table}[!]
\centering
\normalsize
\caption{\textbf{Impact of Text Description Quality on the AIGCDetectBench \cite{zhong2023patchcraft} and AIGIBench \cite{li2025artificial} Dataset.} }
\label{tab:text_description_analysis}
    \begin{tabular}{c|cc}
    \bottomrule
    Models       & AIGCDetectBench      & AIGIBench      \\ \midrule
    ClipCap                 & 91.2      & 82.7           \\
    Qwen-7B                 & 93.2      & 85.7           \\ \midrule
\rowcolor{light-gray} BLIP  & 93.2      & 87.0  \\ \bottomrule
    \end{tabular}
\end{table}

\noindent\textbf{Patch Selection \& Contrastive Learning.} Incorporating only Patch Selection improves the mean Acc. to 82.8\%, a 1.6\% increase over the baseline, indicating that selectively preserving high/low-entropy regions indeed helps retain complementary artifact cues while mitigating the influence of irrelevant semantics beyond standard resizing/cropping. What's more, Contrastive Learning improves detection accuracy by 3\%, indicating that Contrastive Learning effectively encourages the model to learn the distributional differences between real and fake image features, enhancing its discriminative ability. When combining Patch Selection and Contrastive Learning, performance further rises to 87.5\%/80.4\%, showing clear complementarity: Patch Selection improves the quality and diversity of artifact evidence, while Contrastive Learning improves the training signal to emphasize stable structure. 

\subsection{Plug \& Play Application}
To further verify the wide application capability of our method, we apply the NULL-Space Decoupling and Patch Selection to the existing detection framework. They all improve the detection ability of the model, as shown in Table \ref{tab:plug_decoupling} and Table \ref{tab:plug_patch}, suggesting that our method is more effective in promoting the extraction of more generalized artifact information.

\noindent\textbf{NULL-Space Decoupling.} To evaluate the broad applicability of our NULL-Space decoupling module within CLIP-based feature extraction frameworks, we integrated it into the established detection models UnivFD and VIB-Net. As shown in Table \ref{tab:plug_decoupling}, our NULL-Space decoupling module effectively improves detection performance. Specifically, our decoupling module improves the average accuracy by 13.9\% and 2.2\% on UnivFD and VIB-Net, respectively. The larger improvement on UnivFD implies that models dependent on raw CLIP representations are more susceptible to semantic information, whereas VIB-Net has partially mitigated such bias but still benefits from further semantic decoupling. Overall, these indicate that reasonable decoupling of semantic information helps steer the model toward task-relevant forensic cues.

\noindent\textbf{Patch Selection.} We also performed a plug-and-play evaluation on our Patch Selection. The comparison results, illustrated in Table \ref{tab:plug_patch}, demonstrate a consistent improvement in detecting synthetic images from various generative models, including the proposed method of using crops during detection SAFE \cite{li2025improving}, which improves the mean accuracy by 4.8\% on the AIGCDetectBench \cite{zhong2023patchcraft} dataset. The results demonstrate that Patch Selection enables the detector to extract more generalizable features indicative of artifacts, while mitigating the adverse effects of semantic content on artifact detection. This approach enhances the detector’s ability to identify subtle artifacts in input samples and improves its generalization performance.

\begin{figure*}
    \centering
    \includegraphics[width=1\linewidth]{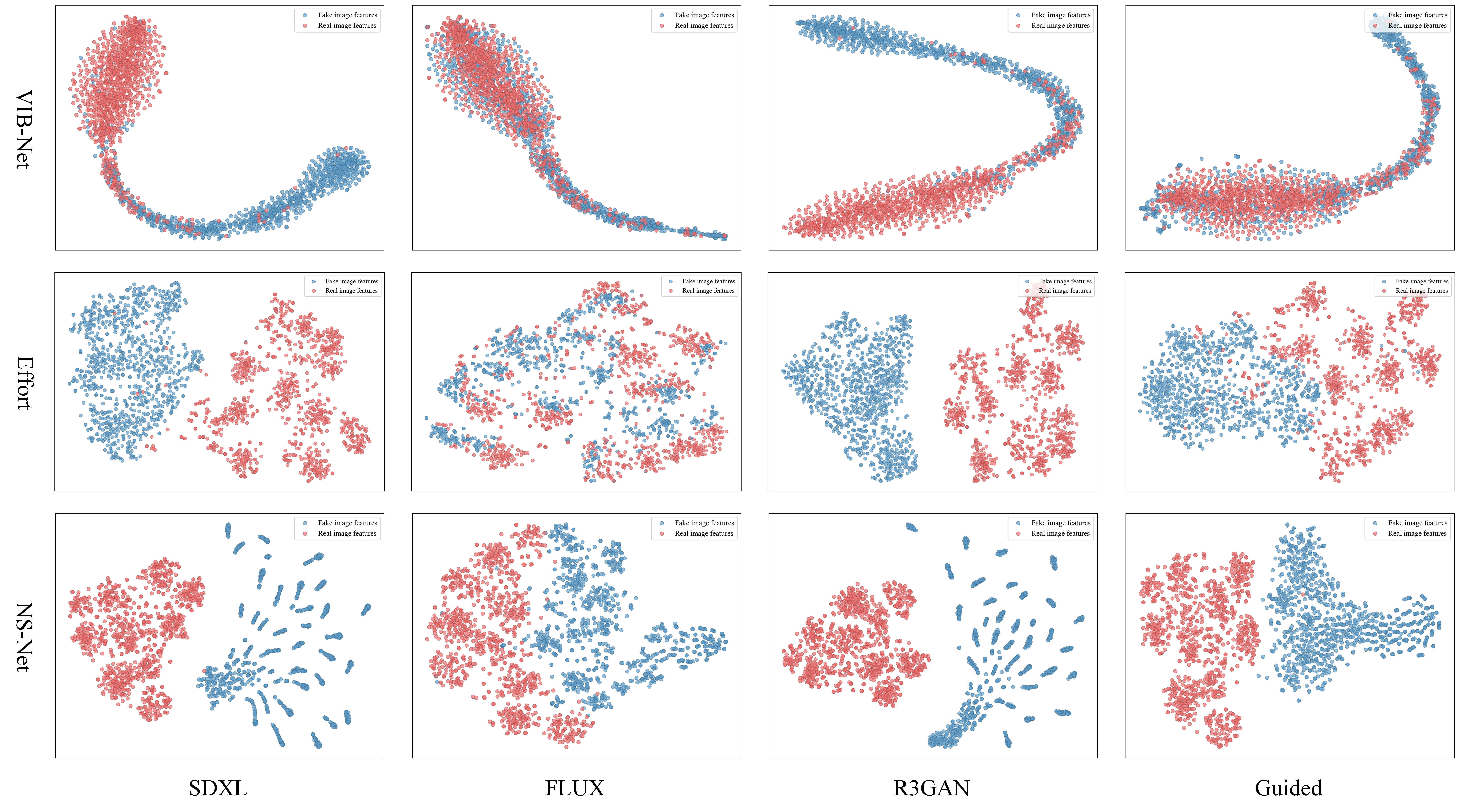}
    \caption{\textbf{T-SNE Visualization of Features Extracted before Classifier}. We compare the VIB-Net, Effort and our NS-Net. A total of four testing GANs and diffusion models are considered, including SDXL, FLUX, R3GAN, and Guided. VIB-Net does not adequately represent semantic content, leading to difficulties when encountering unseen models, while the pre-training knowledge retained by Effort still affects generalization performance to some extent. In contrast, our method demonstrates superior discrimination, effectively separating different categories, even when encountering unseen sources.}
    \label{fig:Qualitative}
\end{figure*}

\subsection{Patch Selection Analysis}
\label{sec:patchselection_analysis}
To further verify the effectiveness of our Patch Selection strategy, we conducted additional experiments, and the results are shown in Table \ref{tab:selection_comparison}. Using Resize or CenterCrop leads to a significant performance drop. PatchShuffle, which applies random shuffling after CenterCrop, achieves comparable performance across the two datasets, suggesting that disrupting global spatial layouts does not cause a substantial degradation in detection performance. Our Patch Selection consistently achieves the best performance across both datasets, indicating that it preserves more discriminative artifact cues.

We further compare RandomSelect, Top-2k high-entropy patches, and Bottom-2k low-entropy patches under the same patch budget. Selecting only high- or low-entropy patches does not yield the best overall performance, whereas the proposed combination of high- and low-entropy patches performs best. This result suggests that high- and low-entropy regions provide complementary artifact cues for cross-model detection.

We also evaluate different patch sizes in Table~\ref{tab:selection_comparison}. Although the performance variation across patch sizes is limited, the $32 \times 32$ setting achieves the best overall performance, providing a suitable balance between preserving local artifact patterns and retaining sufficient local context.

In addition, the most similar to our Patch Selection is PatchCraft \cite{zhong2023patchcraft}, which extracts patches via extensive RandomCrop, and selects texture-rich/poor patches to reconstruct two images. However, this can yield locally redundant patches with highly similar patterns, which may cause reliance on repetitive texture cues and degrade generalization. The result is shown in Table \ref{tab:selection_comparison}, which achieves 77.1\% mAcc. on AIGCDetectBench, supporting the effectiveness of our Patch Selection strategy.

\subsection{Impact of Text Description Quality}
To investigate the dependency of NS-Net on text descriptions, we evaluate the method using captions generated by different image-to-text models, including ClipCap \cite{mokady2021clipcap} and Qwen-7B \cite{qwen}. For each caption generator, we use its generated image descriptions to retrain NS-Net under the same experimental configuration, and evaluate it using the same cross-model generalization protocol. As shown in Table~\ref{tab:text_description_analysis}, NS-Net remains effective with captions generated by different image-to-text models, achieving 91.2 - 93.2\% and 82.7 - 87.0\% mAcc. on AIGCDetectBench and AIGIBench, respectively. These results indicate that NS-Net is not tied to BLIP, although severely inaccurate descriptions may still introduce noisy semantic directions.

\subsection{Robustness Evaluation}
In real-world scenarios, images are inevitably affected by unknown disturbances during transmission and interaction, which pose additional challenges for AI-generated image detection. To investigate robustness under such conditions, we further evaluate the performance of different detection methods against a variety of disturbances, such as JPEG Compression and Gaussian Blur. The results are shown in Table \ref{table:Robustness_AIGCDetectBench} and Table \ref{table:Robustness_AIGIBench}. Experiments on two different datasets reveal that all methods experience performance degradation under perturbations, weakening their ability to reliably distinguish real from fake images. Despite these challenging conditions, our method consistently outperforms others, maintaining a relatively higher accuracy with the mean accuracy of 79.6\% and 66.9\% in QF=75 on two datasets respectively. This indicates that our method has the potential to preserve strong detection performance under previously unseen perturbations.

\subsection{Qualitative Analysis}
To further evaluate the generalization ability of our method, we visualize the feature distribution before classification, as shown in Fig \ref{fig:Qualitative}. This visualization demonstrates how effectively each trained model distinguishes between real and fake images, showcasing our method's ability to generalize across diverse fake representations. From the visualization, it is apparent that, while VIB-Net successfully mitigates some harmful semantic information through information bottlenecks, it does not adequately represent semantic content, leading to difficulties when encountering unseen GAN or diffusion models. As a result, the feature distributions of real and generated images overlap significantly, causing the model to often misclassify fake images as real. 

In addition, Effort, which also uses SVD, improves model generalization by preserving semantic classification knowledge from the pre-trained model. The visualization shows that real and generated images are somewhat clustered, indicating a degree of classification capability, but the distinction between the two categories remains unclear, suggesting limited generalization ability. On the other hand, our method demonstrates superior discrimination, effectively separating different categories, even when encountering unseen sources.

\section{Conclusion}
In this paper, we revisit the distinctions between features with different intrinsic properties extracted by CLIP, and innovatively proposed that the semantic information embedded in the visual features extracted by CLIP's image encoder seriously interferes with the detector's extraction of artifact features. Accordingly, we proposed an innovative method, NS-Net. Specifically, we use the feature homogeneity extracted by the text encoder to replace the semantic information of the features extracted by the image encoder, and use NULL-Space to decouple the semantic information, retaining the artifact information related to the forgery detection task. At the same time, we proposed Patch Selection and used contrastive learning to retain artifact information and promote the model to learn the underlying distributional differences between real and generated images, thereby jointly improving the generalization effect of the model. Extensive experiments on 47 diverse generative models strongly demonstrate the generalization capability of our method.

\bibliographystyle{unsrt}
\bibliography{ieee}

\end{document}